\documentclass[sigconf]{acmart}
\usepackage{url}
\usepackage{subcaption}
\usepackage{graphicx}
\usepackage{amsmath}
\usepackage{amsthm}
\usepackage{booktabs}
\usepackage{algorithm}
\usepackage{algorithmic}
\usepackage[switch]{lineno}

\newtheorem{theorem}{Theorem}
\AtBeginDocument{%
 }

\copyrightyear{2024}
\acmYear{2024}
\setcopyright{rightsretained}
\acmConference[CIKM '24]{Proceedings of the 33rd ACM International Conference on Information and Knowledge Management}{October 21--25, 2024}{Boise, ID, USA}
\acmBooktitle{Proceedings of the 33rd ACM International Conference on Information and Knowledge Management (CIKM '24), October 21--25, 2024, Boise, ID, USA}
\acmDOI{10.1145/3627673.3679575}
\acmISBN{979-8-4007-0436-9/24/10}


\begin{document}

\title{Leveraging Local Structure for Improving Model Explanations: \\ An Information Propagation Approach}

\author{Ruo Yang}
\email{ryang23@hawk.iit.edu}
\affiliation{%
  \institution{Department of Computer Science, Illinois Institute of Technology}
  \city{Chicago}
  \state{Illinois}
  \country{USA}
}
\author{Binghui Wang}
\email{bwang70@iit.edu}
\affiliation{%
  \institution{Department of Computer Science, Illinois Institute of Technology}
  \city{Chicago}
  \state{Illinois}
  \country{USA}
}
\author{Mustafa Bilgic}
\email{mbilgic@iit.edu}
\affiliation{%
  \institution{Department of Computer Science, Illinois Institute of Technology}
  \city{Chicago}
  \state{Illinois}
  \country{USA}
}

\renewcommand{\shortauthors}{Ruo Yang, Binghui Wang, \& Mustafa Bilgic}
\begin{abstract}
  Numerous explanation methods have been recently developed to interpret the decisions made by deep neural network (DNN) models. For image classifiers, these methods typically provide an attribution score to each pixel in the image to quantify its contribution to the prediction. However, most of these explanation methods appropriate attribution scores to pixels \emph{independently}, even though both humans and DNNs make decisions by analyzing a set of closely related pixels simultaneously. Hence, the attribution score of a pixel should be evaluated \emph{jointly} by considering itself and its structurally-similar pixels. We propose a method called IProp, which models each pixel's individual attribution score as a source of explanatory information and explains the image prediction through the dynamic propagation of information across all pixels. To formulate the information propagation, IProp adopts the Markov Reward Process, which guarantees convergence, and the final status indicates the desired pixels' attribution scores. Furthermore, IProp is compatible with \emph{any} existing attribution-based explanation method. Extensive experiments on various explanation methods and DNN models verify that IProp significantly improves them on a variety of interpretability metrics.
\end{abstract}




\maketitle

\section{Introduction}

With the deployment of deep neural network (DNN) models for safety-critical applications such as autonomous driving \cite{Caesar_2020_CVPR,Chen_2015_ICCV} and medical diagnosis \cite{liu2021relational,cuadros2009eyepacs}, explaining the DNN predictions has become a critical component of decision making processes. For humans to trust the decisions of DNNs, performing well on the target task is necessary but not sufficient; the model should also generate explanations that are interpretable by domain experts. There has been a significant amount of research in this area
\cite{lundberg2017unified,sundararajan2020many,jethani2021fastshap,vstrumbelj2014explaining,ribeiro2016should,fong2017interpretable}.
Often, these approaches measure the importance of a pixel as the pixels influence on the decision made by the underlying DNN model. As such, the pixel importance is typically represented by an attribution/saliency map that has the same size as the input image, with each value indicating the importance of the corresponding pixel for the model's decision on that image. 

\begin{figure}[!t]
     \centering
     \begin{subfigure}[b]{\columnwidth}
         \centering
         \includegraphics[width=\textwidth]{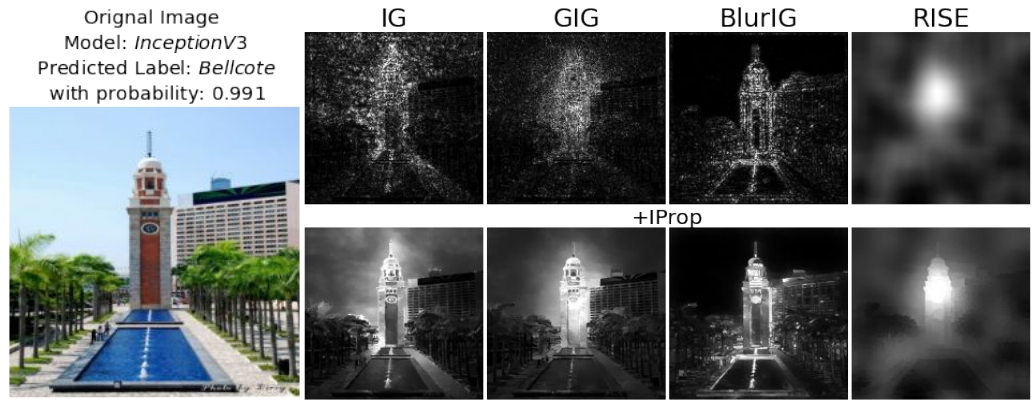}
     \end{subfigure}
\caption{Attribution maps of the existing explanation methods (top row) and those (bottom row) with our information propagation on the \textit{InceptionV3} model. Information propagation ensures maps assign scores more evenly across the object in the image.}
\label{first_graph}
\end{figure}

Most of the current explanation methods construct the attribution map by evaluating the contribution of each pixel \emph{independently}. However, humans and DNNs use the pixels' structural relationships in an image (i.e., locally-connected clusters of pixels) to make predictions. Convolutional neural networks (CNNs), for instance, utilize several layers of convolution and pooling operations to capture the local visual structures in the images. Hence, the maps generated by existing explanation methods are inadequate (See Fig.~\ref{first_graph}). We advocate that modeling and utilizing the local structural relationships between pixels is crucial for designing more effective explanation methods. 
In other words, the attribution scores of pixels should be considered \emph{jointly} for explanation due to pixels' inherent relationships to their neighbor pixels.

One naive strategy to capture pixels' local structure relationships is to first cluster pixels into groups using 
a particular segmentation method and then assign the same attribution score to all pixels in a group, e.g.,  XRAI~\cite{kapishnikov2019xrai}. However, this \emph{static} strategy is suboptimal for several reasons. First, this requires an accurate segmentation approach. Second, even when segmentation is accurate, assuming all pixels in the given segment have the same importance for model decision is a strong assumption.

In this paper, we model pixels' relationships in a \emph{dynamic} way. Specifically, we treat the individual attribution score of each pixel as a source of explanatory information and model the explanation of the prediction of the image to be the dynamic propagation of the individual attribution scores among all pixels of the image.

In this regard, information exchange occurs continously, i.e., information flowing from a pixel to its neighboring pixels and vice versa. Thus, the explanation method \emph{dynamically} measures the information contribution of all pixels. In the ideal situation, such a dynamic process has an equilibrium information distribution in which information exchange ceases. As a consequence of the interaction among pixels, the explanation information for each pixel converges and stabilizes with respect to the information flow. In contrast, if the equilibrium distribution is not achieved, pixels' explanation information exchange continues, indicating the relationships between pixels are not completely exploited. Hence, we endeavor to determine the unique equilibrium information distribution with regard to the dynamic process.

There are two core questions that need to be answered: 
1) How can we model the information flow among pixels?
2) How can we guarantee that the dynamic process converges? 
To address them, we propose an Information Propagation approach (termed IProp) for improving model explanations, that can be applied to the output of \emph{any} existing explanation method that generates an explanation attribution map. Specifically, we first design a weighted graph with pixels as nodes and similarities between pixels as weighted edges, where we investigate the similarity in both the spatial and color space. Next, we model the information propagation among pixels as a Markov Reward Process (MRP), which propagates the pixel's attribution information across nodes (pixels) in the weighted graph, capturing the pixels' structural relationships. We also prove that IProp converges to an unique equilibrium distribution, where each entry's value corresponds to the pixel's final attribution score.   
Finally, we evaluate IProp on multiple explanation metrics with various baseline explanation methods and DNN models for image classification. Our extensive results demonstrate that IProp improves all baselines both qualitatively and quantitatively. 

Our main contributions are summarized as below:
\begin{itemize}
  \item We propose IProp, a novel meta-explanation method, that leverages the local structure relationships of pixels. IProp is compatible with any existing attribution map-based explanation method.

  \item We prove that IProp, which is the dynamic way to model explanation as information propagation among pixels, converges to a unique attribution map when an underlying explanation method is given. 
  
  \item Extensive evaluations show that IProp produces more accurate attribution maps to represent the explanation compared to underlying explanation methods.
\end{itemize}

\begin{figure*}[!ht]
     \centering
     \begin{subfigure}[b]{\textwidth}
         \centering
         \includegraphics[width=\textwidth]{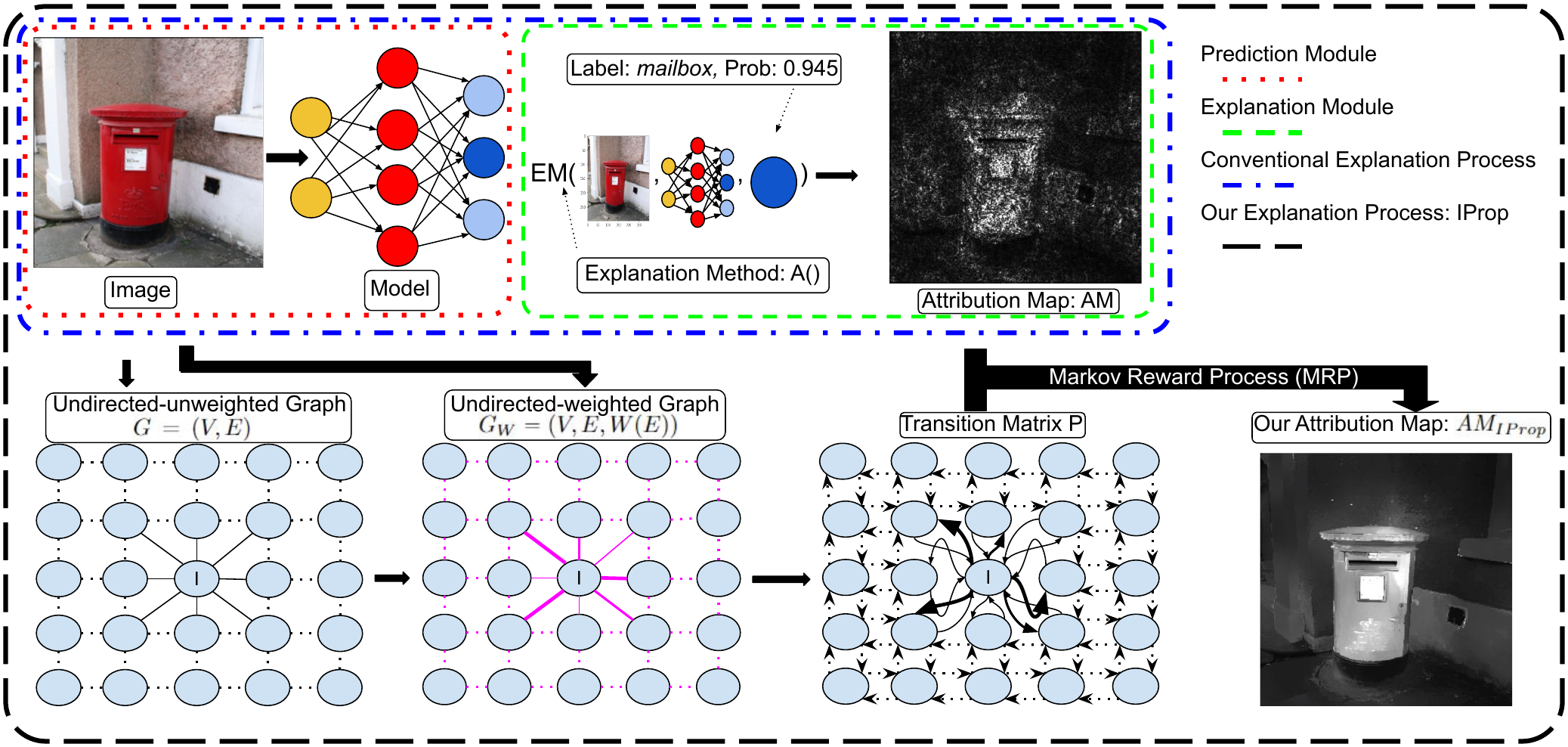}
     \end{subfigure}
\caption{Illustration of IProp. IProp first builds a weighted graph based on  image pixels, where each pixel is a node and the weight of an edge is obtained using the pixels' spatial and color information.  
The weighted graph is associated with a transition matrix. 
Then, IProp performs information propagation based on Markov Reward Process, which takes the transition matrix and pixels' initial rewards as input. Note the pixels' attribution scores (formed as an attribution map), which can be generated by any baseline explanation method, can be treated as the pixels' initial rewards. When the propagation converges, IProp produces pixels' final attribution scores, forming the IProp's attribution map.}
\label{fig_Illustration_of_Our_method}
\end{figure*}

\section{Related Work}
\label{sec_rel}


\noindent {\bf Pixel-based Explanation Methods.} 
Pixel-based explanation methods quantify the contribution of each pixel to the model decision by assigning it an importance score. They can be further categorized as \emph{Shapley value}-, \emph{Input perturbation}-, and \emph{Backpropagation}-based methods. 
The Shapley value~\cite{shapley1953quota} was originally proposed to represent the contribution of each player to the outcome of a cooperative game. 
For explaining image classification, each pixel in an image is treated as a player and the outcome is the image's prediction score. 
Calculating Shapley values exactly is intractable when the image size is large. Hence, several methods propose to approximate the Shapley values, including KernelSHAP~\cite{lundberg2017unified}, BShap~\cite{sundararajan2020many}, and FastShap~\cite{jethani2021fastshap}. \emph{Input perturbation-}based methods work by manipulating the input image and observing its effect on the prediction. This idea is utilized by RISE~\cite{petsiuk2018rise}, the methods learn the mask to use as the attribution maps 
\cite{fong2017interpretable,fong2019understanding}, and other papers~\cite{dabkowski2017real,zintgraf2017visualizing}. \emph{Backpropagation-} based methods propagate the final prediction score back to the input or the hidden layers 
of the DNN and assign a score for each pixel in the input accordingly. These methods include Deconvnet~\cite{zeiler2014visualizing}, guided backpropagation~\cite{springenberg2014striving}, DeepLIFT~\cite{shrikumar2017learning}, LRP~\cite{bach2015pixel}, SmoothGrad~\cite{smilkov2017smoothgrad}, and Grad-CAM~\cite{selvaraju2017grad}. Recently, The Integrated Gradients was proposed by Sundararajan et al.~\shortcite{sundararajan2017axiomatic}. It uses line integration to compute the attribution score for pixels. Its variants include GIG~\cite{kapishnikov2021guided}, Blur IG~\cite{xu2020attribution}, AGI~\cite{pan2021explaining}, and IDGI~\cite{yang2023idgi}.

\noindent {\bf Region-based Explanation Methods.} These types of methods assign attribution scores to each segmented region instead of each pixel. That is, the image is first segmented into distinct regions and the pixels' attribution scores are identical if they are located in the same segment. For example, given an image and an attribution map, XRAI~\cite{kapishnikov2019xrai} creates segments for the image, calculates the attribution score for each segment by summing the attribution scores of all pixels in the segment, and then assigns the same score to all pixels in that segment. Similarly, LIME \cite{ribeiro2016should} first segments the image into superpixels as the features for a linear model, then fits the model where the weights of the model determine the contribution of each superpixel to the prediction. However, the region-based methods do not explicitly consider the structural relationship between pixels, but instead simply assign a score to the pixels based on which segments they belong to.

Our method, IProp, is orthogonal to and compatible with both \emph{region-} and \emph{pixel-}based explanation methods. This is due to the fact that IProp determines the final attribution scores of pixels' by propagating the original attribution scores on a weighted graph (where the weights are determined based on pixel similarities), and the original attribution scores can be obtained via \emph{any} existing explanation method.

\section{Background}

\noindent \textbf{Markov Reward Process (MRP).} MRP models a process where an agent starts in a state, transitions stochastically to a new state based on a probability transition matrix, and 
receives a reward. The discounted cumulative reward~\cite{rao2022foundations} that the agent collects over time $t$ is defined as $G_t = \sum_{i=k+1}^{\infty}\gamma^{i-t-1} \times R_i$, where $\gamma$ is a discounting factor and $R_i$ is the reward at time $i$. $G_t$ can be interpreted as the cumulative reward of a walk on a Markov graph, with each state of the walk contributing the reward $R_i$ with a discounting $\gamma^{i-t-1}$. Then, the {\em value} for a given state $s$, i.e., $V(s)=E[G_t|S=s]$, represents the expected discounted cumulative reward for all paths starting from state $s$ and walking an infinite amount of time. Given the individual reward $R(s)$ for state $s$, the transition matrix $P$ where $P[s,s']$ is the transition probability from state $s$ to $s'$ at any time step, and with the recursion $G_t=R_{k+1}+\gamma \times G_{k+1}$, the Bellman equation~\cite{sutton2018reinforcement} for MRP formally defines the value for the state $s$ as $V(s)=R(s)+\gamma \times \sum_{s' \in S} P[s, s'] \times V(s')$, or in the matrix form as $V=R+\gamma \cdot P\times V$. MRP can be employed to examine the long-term behavior of a system, such as the total reward an agent is expected to accumulate over an infinite number of time steps. 

\noindent \textbf{Model Explanation and Attribution Map.} The aim of an explanation is to determine the importance of the input with respect to the model's prediction.
Given a classifier $f$, class $c$, and an input $x$, let output $f_c(x)$ represent the confidence score (e.g., probability) for predicting $x$ as belonging to class $c$. Formally, the explanation method, $EM()$, is a function that takes the target class $c$, classifier $f$, and input $x$ as input and outputs the attribution map ($AM$), i.e., $AM=EM(f,x,c)$, that has the same size as $x$. Each value $AM_i$ then indicates the importance/attribution score for the $i$-th entry in $x$. In the image classification domain, which is the focus of our paper, $AM$ indicates the attribution scores of all pixels in the image $x$ for a classifier $f$ to make the prediction $f_c(x)$.

\section{IProp: Information Propagation for Improved Model Explanation}
\label{sec_method}

\subsection{Intuition}

\label{intuition}
Almost all the existing explanation methods consider the pixels \emph{independently} when calculating a pixel's contribution to the prediction. However, DNN models make predictions using a collection of structurally-similar pixels rather than using individual ones. This implies that within the context of the model explanation, when assigning an attribution score to a pixel, we should also consider the attribution scores of other structurally similar pixels. One straightforward way to capture the structural similarity is to consider image segmentation to group the pixels. For instance, we can first cluster pixels into segments and assign the \emph{same} attribution score to all the pixels in each segment (similar to XRAI~\cite{kapishnikov2019xrai}). 
However, the output of XRAI depends on the image segmentation technique. For instance, an object may be divided into distinct regions, with pixels from each segment having strong relationships. Then, XRAI assigns different scores to these pixels. Conversely, it is also possible to segment two distinct objects into the same region, in which the pixels in different objects do not share any strong relationships but XRAI assigns the same score for them. 

We propose exploring the inherent relationships between pixels' attribution scores in a \emph{dynamic} way. Specifically, we treat an image as a directed graph where pixels are the nodes and the weights for the directed edges are the nodes' transition probabilities converted from the nodes' similarities. The similarities are computed based on nodes' spatial and color distances. We then model pixels' attribution generation as a dynamic process (i.e., Markov reward process), where each node/pixel's reward is the attribution score from any existing explanation method. Then each pixel is dynamically rewarded during the process which updates its attribution score. Next, we ask if a particle begins at a pixel (e.g., $I$), traverses the weighted graph, and receives the discounted reward from each node along the path of traversal at each time step, what is the expected cumulative attribution reward for the particle after traversing an infinite number of time steps? Importantly, the particle has a larger possibility of visiting structural-similar nodes since large transition probabilities exist between these nodes, which are the normalized similarities. The expected cumulative reward for the particle is treated as the final attribution score of the pixel $I$. 
Now by putting particles on all pixels, such a dynamic process simulates information propagation among all pixels and their structurally-similar counterparts. When the dynamic process converges, we have all pixels' final attribution scores, forming a new attribution map. 

\subsection{The Design of IProp}
\label{sec:design}

Inspired by the above described dynamic information propagation, 
IProp consists of three main steps: 1) Building a weighted graph; 2) Constructing the transition matrix; and 3) Utilizing the Markov Reward Process (MRP) to generate the attribution map. Next, we explain each of the steps in detail. 

\begin{figure}[!t]
     \centering
     \begin{subfigure}[b]{.22\textwidth}
         \centering
         \includegraphics[width=\textwidth]{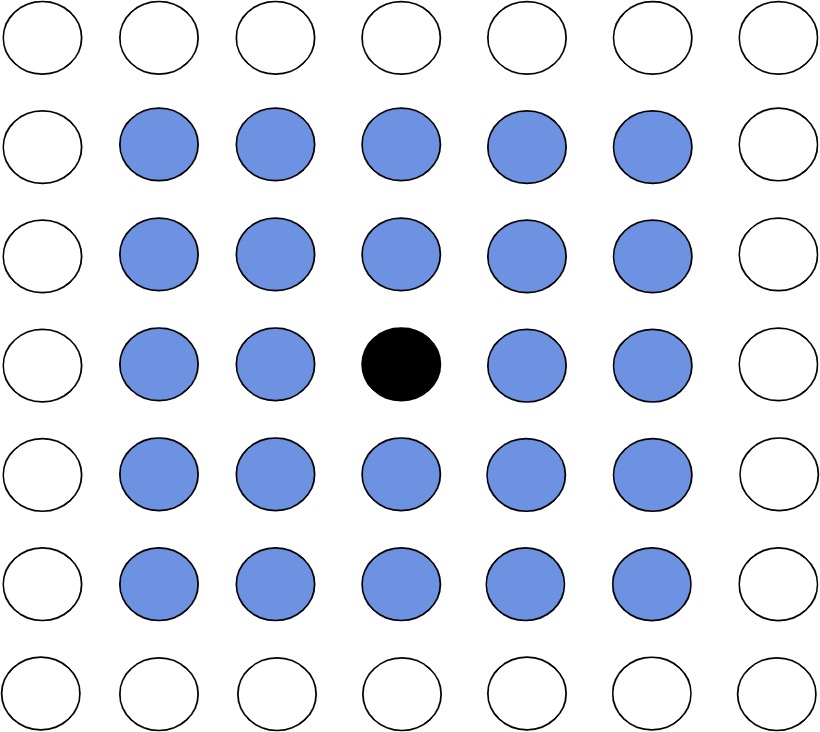}
     \end{subfigure}
     \hfill
     \begin{subfigure}[b]{.22\textwidth}
         \centering
         \includegraphics[width=\textwidth]{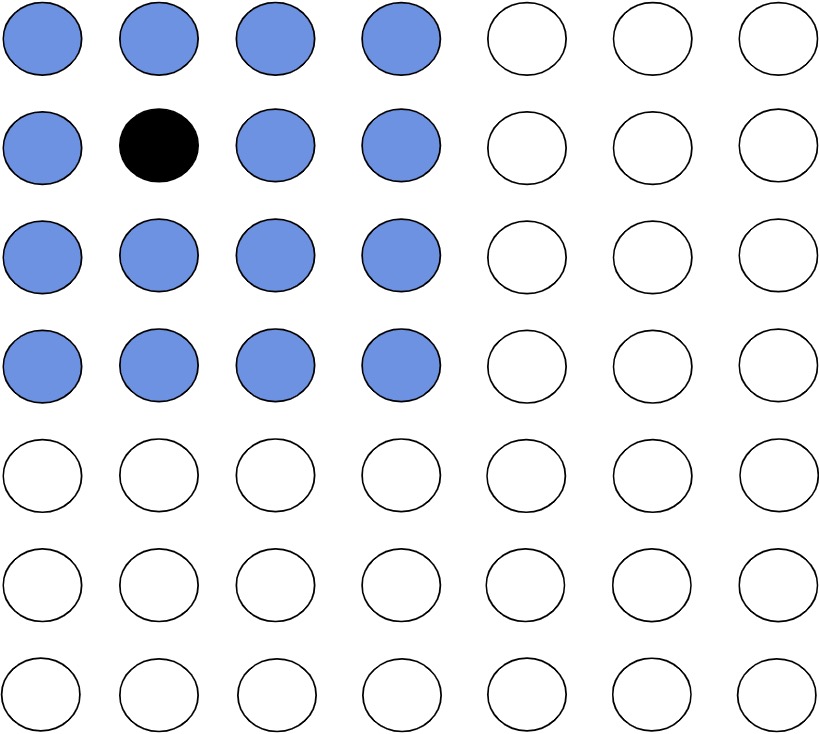}
     \end{subfigure}
\caption{Example of neighboring (blue) nodes for a given (black) node when $K=2$. 
}
\label{fig_buildgraph}
\end{figure}

\noindent \textbf{Building a Weighted Graph.}
Given an image, we treat each pixel as a node. To build the graph, we need to determine the neighborhood of each pixel. For instance, we consider connecting each pixel to its $K$-order neighborhood, where the $K$-order neighborhood pixels have spatial distance K or lower to the target pixel. Hence, each pixel contains at most $(2 \times K+1)^2-1$ neighbors. See Fig.~\ref{fig_buildgraph} for an example when $K=2$. Applying to all pixels, we build an undirected-unweighted graph $G=(V,E)$. Next, we define edge weights. The weight of an edge represents the similarity (or inverse distance) between two connected pixels. There are several methods for measuring such similarity. Here, we are inspired by SLIC~\cite{achanta2012slic,achanta2010slic}, which defines pixel distance as the combination of spatial distance and color distance.
Specifically, the image is first converted to the CIELAB space from the RGB color space. Similar to RGB, each pixel $I$ in the CIELAB space has three values, i.e., $l_I, a_I, b_I$. Then the spatial distance between two pixels $I = (i_I,j_I)$ and $J = (i_J,j_J)$ is defined as the Euclidian distance $d_s^{I,J} = \sqrt{(i_I-i_J)^2+(j_I-j_J)^2}$, and the distance in the CIELAB space is defined as $d_c^{I,J} = \sqrt{(l_I-l_J)^2+(a_I-a_J)^2+(b_I-b_J)^2}$. Finally, the combined distance, i.e., $d^{I,J}=d_c^{I,J}+d_s^{I,J}$, defines the distance between two nodes/pixels. We investigate the ranges of both distances in Section \ref{practial_analysis}. 
Since a longer distance implies less similarity, for simplicity, we define the weight, e.g., $W(I,J)$, between two pixels $I,J$, as their negative distance, i.e., $-d^{I,J}$. 
We denote the undirected weighted graph as $G_W=(V,E, W(E))$. 

\noindent \textbf{Constructing the Transition Matrix.} 
A key step in applying MRP is to first construct the transition matrix, which consists of transition probabilities between two states. 
Intuitively, each node is associated with a state and if two nodes are closer, then the transition probability between these two nodes is larger. Moreover, the transition probabilities from a node to all the other connected nodes sum to 1. 
To capture these intuitions, we propose to convert the weights $W(I,J)$ to probabilities via the softmax function based on the connectivity for node $I$. Specifically, we define the  
transition matrix as $P$ where the $(I, J)$-th entry stands for the similarity between nodes $I$ to $J$. 
Formally,
\begin{equation}\label{softmax}
    P[I,:] = \textrm{softmax}(W[I,:]),
\end{equation}
where $\textrm{softmax}(z)_i = \frac{e^{z_i}}{\sum_{j=1}^{K}e^{e_j}}$. 
Note the transition matrix $P$ is asymmetric since the local structural similarity from $I$ to $J$ is not necessarily the same as that from $J$ to $I$. 

\noindent \textbf{MRP for Generating IProp Attribution Map.}
As mentioned in the Background, MRP determines an equilibrium distribution over all state values by transmitting states' individual rewards according to a predefined transition matrix, such that similar states have similar state values. 
In the context of modeling prediction explanation, we treat the IProp attribution value of a pixel/node as the value of the state/node, e.g., $AM_{IProp}$. Then the initial pixels' attribution scores, e.g., obtained by any existing explanation method, are the pixels' individual rewards. With the MRP, the reward for transitioning from a state/node $j$ to $i$ results a in reward $R$ that is equal to $i$'s initial attribution value, e.g., $AM_i$.

Then, we propagate the information of a pixel's individual reward to other pixels by utilizing the MRP associated with the transition matrix $P$. In this case, the pixel/state reward naturally contains attribution information from other structurally similar pixels. So, the final IProp attribution value of a pixel/node is the value of the state after propagation ends. In other words, for each state/pixel, we start a walk as a player from the state, and the next state of the walk depends on the transition probability (similarity between pixels). Then, a reward is assigned to the player at each step. The final state value represents the expected cumulative reward for the player after walking with infinite steps. Formally, given an initial attribution map $AM$, the discounting factor $\gamma$, and the transition matrix $P$, we obtain the attribution map of IProp, i.e., $AM_{IProp}$, as: 
\begin{equation}
    AM_{IProp} = AM + \gamma \cdot P \cdot AM_{IProp}.
\end{equation}
Directly obtaining the solution $AM_{IProp}$ is computationally challenging, as it needs to solve 
the inverse matrix $(I_N - \gamma P \cdot AM_{IProp})^{-1}$ of size $N$, where $N$ is the number of image pixels that is often large.  In practice, 
since $P$ is highly sparse, we often use the value  iteration method~\cite{rao2022foundations,sutton2018reinforcement,watkins1992q,10.1145/3431379.3460650,kang2022study,sutton2018reinforcement} to iteratively update $AM_{IProp}$. In the $k+1$ iteration, we have:
\begin{equation}
\label{value_ite}
    AM_{IProp}^{k+1} = AM + \gamma \cdot P \cdot AM_{IProp}^{k},
\end{equation}
where $AM_{IProp}^{0}=AM$.
We stop the iteration process until the MSE between $AM_{IProp}$ from two consecutive iterations is smaller than a given tolerance $tol$. 
We also prove the convergence of IProp (Theorem. \ref{theorem_1}) in the appendix.

\begin{theorem}
\label{theorem_1}
The value iteration in IProp (Eq. \ref{value_ite}) is guaranteed to converge to the unique solution $AM_{IProp}^{*}$ for any initial $AM_{IProp}^{0}$, i.e., $\lim_{k \rightarrow \infty}AM_{IProp}^{k} = AM_{IProp}^{*}$. s.t. $AM_{IProp}^{*} = (I_{N} - \gamma P)^{-1} \cdot AM$.
\end{theorem}

\begin{algorithm}[!t]
\footnotesize
    \caption{Pseudo-code for IProp}
    \label{IProp_algo}
    \begin{algorithmic}[1]
    \STATE \textbf{Input:} \text{$x$, $f$, $c$, $EM$, $K$, $\gamma$, $tol$}
    \STATE \textbf{Initialize:} \text{Sets: $V = []$, $E = []$, Matrices: $W = \infty$, $P=0$}
    \STATE \textbf{For} pixel $I$ in $x$ \textbf{do}
    \STATE \quad neighbors($I$) = K-order Neighbor($I$)
    \STATE \quad \textbf{For} pixel $J$ in neighbors($I$) \textbf{do}
    \STATE \quad \quad $V.append(J)$ \textbf{if} $J \notin V$
    \STATE \quad \quad $E.append(edge(I,J))$ \textbf{if} $edge(I,J) \notin E$
    \STATE $l, a, b \equiv x_{CIELAB} = CIELAB(x)$
    \STATE \textbf{For} edge $(I,J) \in E$ \textbf{do}
    \STATE \quad $d_s^{I,J} = \sqrt{(i_I-i_J)^2+(j_I-j_J)^2}$
    \STATE \quad $d_c^{I,J} = \sqrt{(l_I-l_J)^2+(a_I-a_J)^2+(b_I-b_J)^2}$
    \STATE \quad $W[I,J] = W[J,I] = -(d_s^{I,J} + d_c^{I,J})$
    \STATE \textbf{For} row index $I \in W$ \textbf{do}
    \STATE \quad $P[I,:] = softmax(W[I,:])$
    \STATE $AM = EM(x, f, c)$
    \STATE $AM_{IProp}^{old} = AM, AM_{IProp}^{new} = \infty$
    \STATE \textbf{While} MSE($AM_{IProp}^{old}$, $AM_{IProp}^{new}$) $> tol$ \textbf{do}
    \STATE \quad $AM_{IProp}^{new} = AM + \gamma P \cdot AM_{IProp}^{old}$
    \STATE \quad $AM_{IProp}^{old} = AM_{IProp}^{new}$
    \STATE \textbf{Output: $AM_{IProp}^{new}$}
    \end{algorithmic}
\end{algorithm}
\setlength{\textfloatsep}{+2mm}

\section{Experiments}

\subsection{Experimental Setup}
We first generate the attribution map for a model and image using a baseline method (please see below for the baseline methods). Then, we use the IProp to obtain the improved attribution map. We compare the original attribution map and its IProp version both qualitatively and quantitatively.

\noindent \textbf{Baselines.}  We use eight \emph{pixel-} and \emph{region-}based explanation methods as the baselines. For \emph{pixel-}based explanation methods, we consider Integrated Gradients (IG) \cite{sundararajan2017axiomatic}, GIG~\cite{kapishnikov2021guided}, and BlurIG~\cite{xu2020attribution} as the IG-based methods. We follow previous work~\cite{kapishnikov2021guided} to set the black image as the reference point for IG and GIG, use a step size of 200 as the parameter, and utilize the original implementations with default parameters in the authors' code for all three IG-based methods. We also include the Vanilla Gradient (VG)~\cite{simonyan2013deep}, and follow the original settings for the RISE \cite{petsiuk2018rise} which generates 4K $7\times7$ binary masks first and then upsampling to the original image size for computing the attribution map for each image. Lastly, we include the Grad-CAM (GCAM) \cite{selvaraju2017grad} with the activations from the last CNN layers.

For \emph{region-}based explanation methods, we implement  LIME~\cite{ribeiro2016should}, which works as a superpixel-based explanation method in the image domain. For each image, we first utilize SLIC~\cite{achanta2012slic,achanta2010slic} to segment the image into 200 superpixels (regions) and then generate 4K random binary masks of size 200 with equal probability to be 1 or 0. 1 indicates the superpixel is turned on while 0 means the superpixel is turned off (replace the superpixel with black values). Then we train a logistic regression and report the weights as the importance for the superpixels. Similar to LIME, we modify the original RISE by replacing the randomly generated mask from the pixel level with the superpixel level and utilize the original RISE mechanism to compute the importance score for each superpixel. As the same setting for generating the binary masks, we use 4K samples for each image and refer to the modified version as RISE(S) for the superpixel level. Furthermore, we use the original authors' implementation\footnote{https://github.com/PAIR-code/saliency} of XRAI as another region-based explanation baseline approach.



\begin{figure}[!t]
     \centering
     \begin{subfigure}[b]{\columnwidth}
         \centering
         \includegraphics[width=\textwidth]{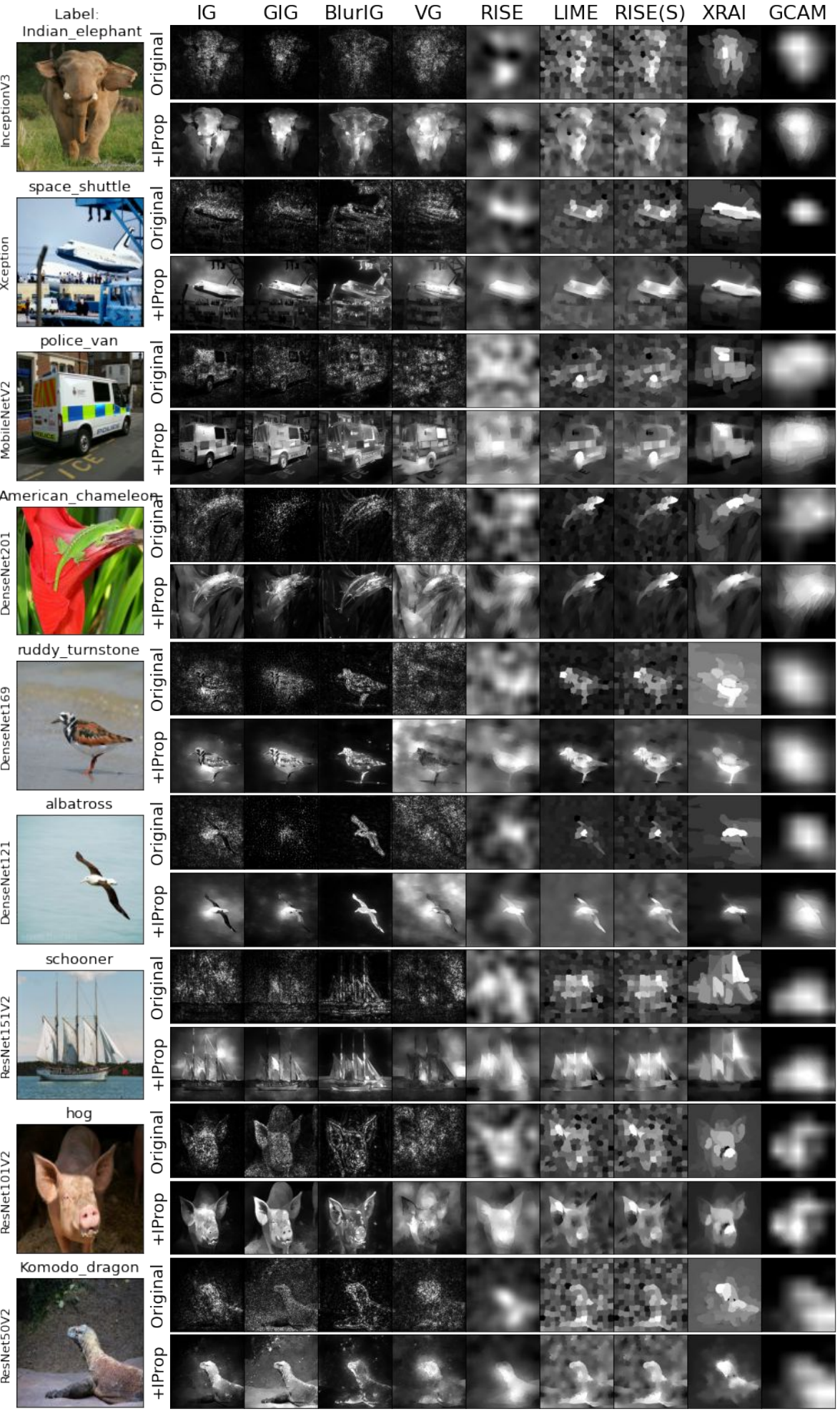}
     \end{subfigure}
\caption{Attribution map of baseline methods and that with IProp. IProp ensures attribution maps focus more on the object,  while baseline methods assign attribute scores to many pixels not in the object.}
\label{Qualitative}
\end{figure}

\noindent \textbf{Models and Running Machine.} We use nine well-known TensorFlow(2.3.0)~\cite{tensorflow2015-whitepaper} 
pre-trained image classifiers: DenseNet\{121, 169, 201\} \cite{huang2017densely}, InceptionV3 \cite{szegedy2016rethinking}, MobileNetV2 \cite{howard2017mobilenets}, ResNet\{50, 101, 152\}V2 \cite{he2016deep}, and Xception \cite{chollet2017xception}. We use a machine with dual Intel(R) Xeon(R) Silver 4214 CPUs (24 cores in total), 64G RAM, and two RTX-5000 GPUs for all the experiments.

\noindent \textbf{Testing Set.} Following \cite{kapishnikov2021guided,xu2020attribution,kapishnikov2019xrai,pan2021explaining,qi2019visualizing}, we use the Imagenet \cite{deng2009imagenet} validation dataset, which contains 50K test samples with labels and annotations. We first identify the set of images that are correctly predicted with respect to each model, then we sample 5K images as test instances that need to be explained for that model.

\noindent \textbf{Hyperparameters} By default, we set $K=\text{image size}/32, \gamma=0.99,$ and $tol=1e^{-7}$. We also study the impact of these hyperparameters in later sections.

\noindent \textbf{Evaluation metrics.} We use numerous metrics to quantitatively evaluate our method and the baselines, i.e., Insertion score and Deletion scores~\cite{pan2021explaining,petsiuk2018rise}, Softmax information curves (SIC), Accuracy information curves (AIC) \cite{kapishnikov2019xrai,kapishnikov2021guided}, and ROC-AUC \cite{cong2018review,xu2020attribution,kapishnikov2019xrai}. 
We provide the implementation details in the appendix. 
We will also open-source these implementations.

\begin{table*}[!t]
\centering
\resizebox{1.45\columnwidth}{!}{
\begin{tabular}{|c||c||c|c|c|c|c|c||c|c|c|}
\hline
\textbf{Model}&&\multicolumn{9}{|c|}{\textbf{Explanation methods} ($\uparrow$)}\\
\hline
&&\multicolumn{6}{|c|}{\textbf{Pixel-based methods}}&\multicolumn{3}{|c|}{\textbf{Region-based methods}}\\
\cline{3-11}
&&IG&GIG&BlurIG&VG&RISE&GCAM&LIME&RISE(S)&XRAI\\
\hline

\textit{InceptionV3}&Original&.203&.187&.263&.126&.482&.843&.554&.545&.477\\
&+IProp&\textbf{.451}&\textbf{.451}&\textbf{.479}&\textbf{.435}&\textbf{.522}&\textbf{.872}&\textbf{.570}&\textbf{.567}&\textbf{.492}\\
\hline
\textit{Xception}&Original&.222&.225&.294&.159&.492&.859&.584&.572&.486\\
&+IProp&\textbf{.483}&\textbf{.497}&\textbf{.505}&\textbf{.478}&\textbf{.530}&\textbf{.884}&\textbf{.591}&\textbf{.585}&\textbf{.510}\\
\hline
\textit{MobileNetV2}&Original&.099&.119&.150&.070&.452&.771&.516&.506&.407\\
&+IProp&\textbf{.383}&\textbf{.419}&\textbf{.408}&\textbf{.377}&\textbf{.495}&\textbf{.805}&\textbf{.528}&\textbf{.525}&\textbf{.432}\\
\hline
\textit{DenseNet201}&Original&.173&.167&.204&.103&.468&.828&.549&.540&.439\\
&+IProp&\textbf{.425}&\textbf{.450}&\textbf{.442}&\textbf{.404}&\textbf{.525}&\textbf{.863}&\textbf{.570}&\textbf{.563}&\textbf{.478}\\
\hline
\textit{DenseNet169}&Original&.177&.164&.193&.097&.485&.821&.568&.561&.468\\
&+IProp&\textbf{.453}&\textbf{.470}&\textbf{.463}&\textbf{.422}&\textbf{.531}&\textbf{.847}&\textbf{.581}&\textbf{.576}&\textbf{.497}\\
\hline
\textit{DenseNet121}&Original&.155&.146&.183&.086&.466&.809&.558&.551&.438\\
&+IProp&\textbf{.433}&\textbf{.440}&\textbf{.456}&\textbf{.394}&\textbf{.522}&\textbf{.849}&\textbf{.575}&\textbf{.570}&\textbf{.480}\\
\hline
\textit{ResNet152V2}&Original&.182&.164&.201&.111&.440&.697&.472&.468&.411\\
&+IProp&\textbf{.405}&\textbf{.413}&\textbf{.430}&\textbf{.379}&\textbf{.489}&\textbf{.723}&\textbf{.500}&\textbf{.498}&\textbf{.437}\\
\hline
\textit{ResNet101V2}&Original&.175&.165&.198&.111&.443&.711&.486&.482&.415\\
&+IProp&\textbf{.398}&\textbf{.414}&\textbf{.436}&\textbf{.388}&\textbf{.488}&\textbf{.740}&\textbf{.509}&\textbf{.506}&\textbf{.437}\\
\hline
\textit{ResNet50V2}&Original&.168&.169&.196&.115&.449&.711&.478&.473&.402\\
&+IProp&\textbf{.389}&\textbf{.414}&\textbf{.430}&\textbf{.389}&\textbf{.499}&\textbf{.746}&\textbf{.498}&\textbf{.496}&\textbf{.427}\\
\hline

\end{tabular}}
\caption{AUC for AIC. IProp improves all baselines.}
\label{aic}
\end{table*}

\subsection{Qualitative Results}

Figure \ref{Qualitative} visualizes the attribution map (of a set of randomly chosen images for each of the 9 models) obtained by the baseline explanation methods, and the its IProp version.
We observe that the baseline attribution maps may not focus on the object itself, and contain noisy attribution scores outside the object. After applying IProp, however, the attribution map has more uniform attribution scores for the objects' relevant pixels. This shows that information propagation can capture pixels' local structural relationships, and hence can help to better explain the predictions. However, qualitative and visual inspections are often subjective, thus we focus on the quantitative metrics in the rest of the experiments.


\begin{table*}[!t]
\centering
\resizebox{1.45\columnwidth}{!}{
\begin{tabular}{|c||c||c|c|c|c|c|c||c|c|c|}
\hline
\textbf{Model}&&\multicolumn{9}{|c|}{\textbf{Explanation methods} ($\uparrow$)}\\
\hline
&&\multicolumn{6}{|c|}{\textbf{Pixel-based methods}}&\multicolumn{3}{|c|}{\textbf{Region-based methods}}\\
\cline{3-11}
&&IG&GIG&BlurIG&VG&RISE&GCAM&LIME&RISE(S)&XRAI\\
\hline

\textit{InceptionV3}&Original&.086&.059&.166&.029&.456&.804&.556&.543&.450\\
&+IProp&\textbf{.432}&\textbf{.423}&\textbf{.462}&\textbf{.408}&\textbf{.501}&\textbf{.837}&\textbf{.580}&\textbf{.574}&\textbf{.471}\\
\hline
\textit{Xception}&Original&.109&.099&.207&.048&.461&.816&.573&.556&.458\\
&+IProp&\textbf{.462}&\textbf{.478}&\textbf{.495}&\textbf{.451}&\textbf{.509}&\textbf{.845}&\textbf{.591}&\textbf{.583}&\textbf{.488}\\
\hline
\textit{MobileNetV2}&Original&.020&.023&.045&.011&.415&.708&.493&.477&.351\\
&+IProp&\textbf{.334}&\textbf{.375}&\textbf{.357}&\textbf{.325}&\textbf{.462}&\textbf{.746}&\textbf{.516}&\textbf{.510}&\textbf{.381}\\
\hline
\textit{DenseNet201}&Original&.063&.056&.112&.018&.454&.797&.557&.539&.427\\
&+IProp&\textbf{.410}&\textbf{.437}&\textbf{.425}&\textbf{.376}&\textbf{.517}&\textbf{.835}&\textbf{.582}&\textbf{.571}&\textbf{.467}\\
\hline
\textit{DenseNet169}&Original&.069&.052&.097&.017&.453&.796&.569&.554&.450\\
&+IProp&\textbf{.427}&\textbf{.447}&\textbf{.436}&\textbf{.374}&\textbf{.508}&\textbf{.824}&\textbf{.578}&\textbf{.570}&\textbf{.482}\\
\hline
\textit{DenseNet121}&Original&.046&.034&.080&.014&.446&.775&.549&.533&.407\\
&+IProp&\textbf{.397}&\textbf{.415}&\textbf{.426}&\textbf{.361}&\textbf{.509}&\textbf{.818}&\textbf{.577}&\textbf{.568}&\textbf{.452}\\
\hline
\textit{ResNet152V2}&Original&.095&.063&.119&.024&.443&.679&.497&.486&.414\\
&+IProp&\textbf{.409}&\textbf{.412}&\textbf{.432}&\textbf{.379}&\textbf{.496}&\textbf{.710}&\textbf{.536}&\textbf{.533}&\textbf{.442}\\
\hline
\textit{ResNet101V2}&Original&.094&.073&.117&.026&.456&.698&.515&.504&.424\\
&+IProp&\textbf{.407}&\textbf{.418}&\textbf{.446}&\textbf{.386}&\textbf{.507}&\textbf{.729}&\textbf{.552}&\textbf{.547}&\textbf{.448}\\
\hline
\textit{ResNet50V2}&Original&.084&.072&.108&.026&.452&.693&.497&.489&.401\\
&+IProp&\textbf{.384}&\textbf{.411}&\textbf{.430}&\textbf{.383}&\textbf{.501}&\textbf{.731}&\textbf{.532}&\textbf{.527}&\textbf{.424}\\
\hline
\end{tabular}}

\caption{AUC for SIC. IProp improves all baselines.}
\label{sic}

\end{table*}

\subsection{Quantitative Results}
\label{sec:quan_res}
\noindent {\bf Results on AIC and SIC.}
This evaluation method begins with a blurred version of the target test image and restores the pixels' values of the most important pixels, as decided by the explanation method, resulting in a bokeh image. Then, an information level is calculated for each bokeh image by comparing the size of the compressed bokeh image and the size of the compressed original image. The information level is referred to as the Normalized Entropy. Based on the amount of information, bokeh images are binned. The average accuracy is then calculated for each bin. AIC represents the curve of these mean accuracy across bins. Additionally, the predicted probability of bokeh versus the original image is calculated for each image within each bin. SIC is the curve of the median value over each bin. The areas under the AIC and SIC curves are computed; better explanation methods are expected to have greater values. In Tables \ref{aic} and \ref{sic}, we present the AUC under the AIC and SIC curves for all baselines and with IProp. We observe IProp consistently improves all baselines, suggesting the IProp's explanations are better aligned with what the models do for their predictions.

\begin{table*}[!t]
\centering
\resizebox{1.45\columnwidth}{!}{
\begin{tabular}{|c||c||c|c|c|c|c|c||c|c|c|}
\hline
{\textbf{Model}} &&\multicolumn{9}{|c|}{\textbf{Explanation methods} ($\uparrow$)}\\
\hline
&&\multicolumn{6}{|c|}{\textbf{Pixel-based methods}}&\multicolumn{3}{|c|}{\textbf{Region-based methods}}\\
\cline{3-11}
&&IG&GIG&BlurIG&VG&RISE&GCAM&LIME&RISE(S)&XRAI\\
\hline

\textit{InceptionV3}&Original&.679&.663&.694&.660&.724&.857&.688&.672&.782\\
&+IProp&\textbf{.745}&\textbf{.730}&\textbf{.791}&\textbf{.755}&\textbf{.729}&.857&\textbf{.718}&\textbf{.702}&\textbf{.892}\\
\hline
\textit{Xception}&Original&.694&.702&.706&.682&.718&.866&.695&.673&.791\\
&+IProp&\textbf{.764}&\textbf{.779}&\textbf{.797}&\textbf{.789}&\textbf{.722}&.866&\textbf{.727}&\textbf{.706}&\textbf{.802}\\
\hline
\textit{MobileNetV2}&Original&.677&.695&.684&.652&.738&.823&.687&.675&.789\\
&+IProp&\textbf{.747}&\textbf{.794}&\textbf{.785}&\textbf{.754}&\textbf{.743}&\textbf{.824}&\textbf{.717}&\textbf{.708}&\textbf{.800}\\
\hline
\textit{DenseNet201}&Original&.653&.661&.655&.605&.736&.810&.685&.668&.758\\
&+IProp&\textbf{.712}&\textbf{.751}&\textbf{.754}&\textbf{.675}&\textbf{.742}&\textbf{.811}&\textbf{.719}&\textbf{.703}&\textbf{.769}\\
\hline
\textit{DenseNet169}&Original&.657&.655&.656&.583&.707&.801&.687&.670&.758\\
&+IProp&\textbf{.719}&\textbf{.737}&\textbf{.755}&\textbf{.634}&\textbf{.713}&.801&\textbf{.721}&\textbf{.705}&\textbf{.769}\\
\hline
\textit{DenseNet121}&Original&.663&.661&.662&.609&.712&.803&.687&.671&.751\\
&+IProp&\textbf{.728}&\textbf{.753}&\textbf{.766}&\textbf{.681}&\textbf{.718}&\textbf{.804}&\textbf{.720}&\textbf{.706}&\textbf{761}\\
\hline
\textit{ResNet152V2}&Original&.706&.682&.686&.660&.739&.721&.666&.656&.793\\
&+IProp&\textbf{.762}&\textbf{.761}&\textbf{.792}&\textbf{.741}&\textbf{.745}&\textbf{.721}&\textbf{.696}&\textbf{.687}&\textbf{.804}\\
\hline
\textit{ResNet101V2}&Original&.709&.694&.695&.670&.748&.739&.678&.667&.797\\
&+IProp&\textbf{.764}&\textbf{.770}&\textbf{.803}&\textbf{.754}&\textbf{.754}&\textbf{.740}&\textbf{.710}&\textbf{.701}&\textbf{.809}\\
\hline
\textit{ResNet50V2}&Original&.699&.699&.689&.672&.781&.758&.674&.664&.782\\
&+IProp&\textbf{.749}&\textbf{.775}&\textbf{.798}&\textbf{.757}&\textbf{.788}&\textbf{.759}&\textbf{.705}&\textbf{.697}&\textbf{.793}\\
\hline
\end{tabular}}

\caption{ROC-AUC. IProp improves 78 out of 81 baselines.}
\label{rocauc}

\end{table*}

\noindent {\bf Results on Deletion and Insertion Ratio.}
Further, we evaluate all explanation methods using the Insertion Score and Deletion Score from prior research \cite{pan2021explaining,petsiuk2018rise,qi2019visualizing}. For each test image, the insertion technique inserts pixels, from the highest to lowest attribution score, to the black image, then makes the prediction on the modified image. The method produces a curve that represents the predicted values as a function of the percentage of the number of pixels inputted. In contrast, the deletion method deletes the pixels from the original image by replacing those pixels' values with zeros (black image). The insertion and deletion scores are then determined as the AUC. The higher the insertion score or the lower the deletion score implies the explanation method produces better attribution maps. As indicated in the previous research \cite{qi2019visualizing}, one should consider the insertion and deletion scores jointly. Here, 
we compute the Deletion-Insertion Ratio. The range for both Insertion and Deletion scores is from 0 to 1, and since a better explanation method should give a higher insertion score and a lower deletion score for each image, then the Deletion-Insertion ratio, e.g., $\text{Deletion-Insertion Ratio} = \frac{\text{Deletion Score}}{\text{Insertion Score}}$ should have a lower value for a better explanation method. We report the average Deletion-Insertion Ratio (DIR) over all test images in Table \ref{deletionratio}, and the information propagation improves (decreases) the DIR score from most (72 out of 81) of these baselines. We include Insertion and Deletion scores separately 
in the appendix.



\begin{table*}[!t]
\centering
\resizebox{1.45\columnwidth}{!}{
\begin{tabular}{|c||c||c|c|c|c|c|c||c|c|c|}
\hline
\textbf{Model}&&\multicolumn{9}{|c|}{\textbf{Explanation methods} ($\downarrow$)}\\
\hline
&&\multicolumn{6}{|c|}{\textbf{Pixel-based methods}}&\multicolumn{3}{|c|}{\textbf{Region-based methods}}\\
\cline{3-11}
&&IG&GIG&BlurIG&VG&RISE&GCAM&LIME&RISE(S)&XRAI\\
\hline

\textit{InceptionV3}&Original&.386&\textbf{.341}&.483&.820&.314&.155&.242&.254&.283\\
&+IProp&\textbf{.261}&.379&\textbf{.252}&\textbf{.631}&\textbf{.290}&\textbf{.149}&\textbf{.240}&\textbf{.251}&\textbf{.263}\\
\hline
\textit{Xception}&Original&.398&.286&.431&.763&.351&.175&\textbf{.210}&.227&.273\\
&+IProp&\textbf{.251}&\textbf{.276}&\textbf{.239}&\textbf{.512}&\textbf{.332}&\textbf{.153}&.211&\textbf{.221}&\textbf{.254}\\
\hline
\textit{MobileNetV2}&Original&.546&.355&.487&.974&.255&\textbf{.174}&.254&.268&.305\\
&+IProp&\textbf{.297}&\textbf{.270}&\textbf{.264}&\textbf{.653}&\textbf{.245}&.187&\textbf{.242}&\textbf{.251}&\textbf{.302}\\
\hline
\textit{DenseNet201}&Original&.471&.324&.499&1.005&.307&\textbf{.236}&.238&.258&.357\\
&+IProp&\textbf{.308}&\textbf{.297}&\textbf{.280}&\textbf{.847}&\textbf{.293}&.242&\textbf{.226}&\textbf{.241}&\textbf{.337}\\
\hline
\textit{DenseNet169}&Original&.435&.321&.454&1.117&\textbf{.326}&\textbf{.254}&.224&.249&.346\\
&+IProp&\textbf{.299}&\textbf{.300}&\textbf{.281}&\textbf{.964}&.340&.258&\textbf{.220}&\textbf{.237}&\textbf{.329}\\
\hline
\textit{DenseNet121}&Original&.441&.338&.464&.984&.283&.225&.228&.247&.383\\
&+IProp&\textbf{.292}&\textbf{.281}&\textbf{.261}&\textbf{.787}&\textbf{.273}&\textbf{.224}&\textbf{.221}&\textbf{.232}&\textbf{.356}\\
\hline
\textit{ResNet152V2}&Original&.354&.309&.521&.802&.289&\textbf{.738}&\textbf{.348}&.358&.283\\
&+IProp&\textbf{.249}&\textbf{.296}&\textbf{.274}&\textbf{.627}&\textbf{.278}&.781&.350&\textbf{.356}&\textbf{.258}\\
\hline
\textit{ResNet101V2}&Original&.363&.303&.505&.740&.277&\textbf{.681}&.320&.334&.278\\
&+IProp&\textbf{.245}&\textbf{.299}&\textbf{.277}&\textbf{.572}&\textbf{.274}&.692&\textbf{.311}&\textbf{.315}&\textbf{.271}\\
\hline
\textit{ResNet50V2}&Original&.380&.290&.512&.746&.241&.511&.327&.333&.293\\
&+IProp&\textbf{.260}&\textbf{.266}&\textbf{.263}&\textbf{.574}&\textbf{.240}&\textbf{.491}&\textbf{.319}&\textbf{.329}&\textbf{.278}\\
\hline
\end{tabular}}

\caption{Deletion-Insertion Ratio (DIR) Score. 
72 out of 81 baselines with our IProp have lower DIR scores.}
\label{deletionratio}
\end{table*}

\noindent {\bf Results on ROC-AUC.}
Following 
\cite{cong2018review,xu2020attribution,kapishnikov2019xrai},
this evaluation metric computes the ROC-AUC by considering the attribution values as the prediction scores which determine whether the important pixels are predicted to be inside a given annotation area. This metric measures how the generated attribution map is similar to the human perspective on that image. Note that it does not directly measure the quality of explanation, since the model could have a different ``perspective'' than humans, e.g., focusing on the different regions to make predictions. We report the ROC-AUC results in Table \ref{rocauc}; IProp outperforms most of the baselines.

\noindent {\bf Results on Pointing Game.} The metric \cite{zhang2018top} first finds the pixel with the maximum value in the saliency map, then checks whether the pixel lies in the ground truth annotation provided by humans. In other words, the metric computes a hit rate for each attribution method over all of the test images. Tab. \ref{tab.point} shows the Pointing game scores for all models with different attribution methods. Our method improves the metric from most (52 out of 81) of the baselines.

\begin{table*}[!t]
\centering
\resizebox{1.45\columnwidth}{!}{
\begin{tabular}{|c||c||c|c|c|c|c|c||c|c|c|}
\hline
\textbf{Model}&&\multicolumn{9}{|c|}{\textbf{Explanation methods} ($\downarrow$)}\\
\hline
&&\multicolumn{6}{|c|}{\textbf{Pixel-based methods}}&\multicolumn{3}{|c|}{\textbf{Region-based methods}}\\
\cline{3-11}
&&IG&GIG&BlurIG&VG&RISE&GCAM&LIME&RISE(S)&XRAI\\
\hline

\textit{InceptionV3}&Original&.398&.245&.335&.292&\textbf{.907}&\textbf{.939}&\textbf{.923}&\textbf{.919}&.844\\
&+IProp&\textbf{.641}&\textbf{.653}&\textbf{.552}&\textbf{.715}&.860&.937&.901&.894&\textbf{.871}\\
\hline
\textit{Xception}&Original&.428&.252&.338&.330&\textbf{.899}&\textbf{.946}&\textbf{.931}&\textbf{.925}&.853\\
&+IProp&\textbf{.665}&\textbf{.689}&\textbf{.558}&\textbf{.763}&.883&.941&.905&.897&\textbf{.872}\\
\hline
\textit{MobileNetV2}&Original&.455&.269&.347&.360&\textbf{.947}&\textbf{.852}&\textbf{.931}&\textbf{.931}&.820\\
&+IProp&\textbf{.736}&\textbf{.716}&\textbf{.550}&\textbf{.772}&.918&.849&.906&.900&\textbf{.859}\\
\hline
\textit{DenseNet201}&Original&.404&.196&.247&.263&\textbf{.938}&.869&\textbf{.927}&\textbf{.924}&.813\\
&+IProp&\textbf{.621}&\textbf{.617}&\textbf{.513}&\textbf{.618}&.913&\textbf{.875}&.904&.895&\textbf{.856}\\
\hline
\textit{DenseNet169}&Original&.414&.190&.253&.221&\textbf{.904}&.869&\textbf{.933}&\textbf{.927}&.825\\
&+IProp&\textbf{.642}&\textbf{.604}&\textbf{.537}&\textbf{.557}&.892&\textbf{.873}&.906&.899&\textbf{.860}\\
\hline
\textit{DenseNet121}&Original&.398&.198&.262&.264&.878&.869&\textbf{.928}&\textbf{.923}&.790\\
&+IProp&\textbf{.624}&\textbf{.615}&\textbf{.516}&\textbf{.630}&\textbf{.898}&\textbf{.880}&.897&.893&\textbf{.834}\\
\hline
\textit{ResNet152V2}&Original&.525&.258&.354&.366&\textbf{.927}&.790&\textbf{.921}&\textbf{.923}&.846\\
&+IProp&\textbf{.728}&\textbf{.683}&\textbf{.587}&\textbf{.713}&.918&\textbf{.822}&.900&.899&\textbf{.881}\\
\hline
\textit{ResNet101V2}&Original&.492&.270&.342&.341&\textbf{.917}&.826&\textbf{.927}&\textbf{.930}&.842\\
&+IProp&\textbf{.681}&\textbf{.678}&\textbf{.571}&\textbf{.693}&.903&\textbf{.849}&.901&.905&\textbf{.875}\\
\hline
\textit{ResNet50V2}&Original&.467&.265&.319&.353&\textbf{.944}&.860&\textbf{.930}&\textbf{.929}&.827\\
&+IProp&\textbf{.691}&\textbf{.692}&\textbf{.561}&\textbf{.696}&.933&\textbf{.885}&.905&.905&\textbf{.866}\\
\hline
\end{tabular}}

\caption{Pointing game scores. 52 out of 81 baselines with our IProp have higher Pointing game scores.}
\label{tab.point}
\end{table*}

\noindent {\bf Results on Sanity Checks.} An attribution method should pass sanity checks \cite{adebayo2018sanity}. 
When the base attribution method passes the sanity checks, it produces distinct $AM$s based on different sanity checks. IProp generates distinct $AM$s, as seen in Fig. \ref{Qualitative}. 
Furthermore, following \cite{adebayo2018sanity}, we compare the Spearman rank correlation between the absolute values of the AMs of the pixels, generated using the original model and a model with random weights. Table~\ref{tab.sanity_check} shows that as long as the base attribution method has a low coef, IProp also has a low coef. As expected, IProp slightly increases the base coef, possibly due to the correlations introduced through the neighborhood; however, the IProp coefs still remain small. Adebayo et al.~\shortcite{adebayo2018sanity} showed that IG had $0.5$ and GBP had close to $1$ coef. Hence, IProp is expected to pass the sanity checks as long as the underlying attribution maps satisfy the sanity checks.

\begin{table}[!t]
\centering
\footnotesize
\resizebox{\columnwidth}{!}{
\begin{tabular}{|c|c|||c|c|||c|c|||c|c|}
\hline

\multicolumn{8}{|c|}{\textbf{Explanation methods}}\\
\hline
IG&+IProp&GIG&+IProp&BlurIG&+IProp&VG&+IProp\\
\hline
.476&.672&.302&.470&.295&.411&.215&.342\\
\hline

\end{tabular}}
\caption{Spearman rank correlation for the Sanity check with model parameter randomization test on \textit{InceptionV3} model.}
\label{tab.sanity_check}
\end{table}

\begin{figure}[!t]
     \centering
         \centering
\includegraphics[width=0.48\textwidth]{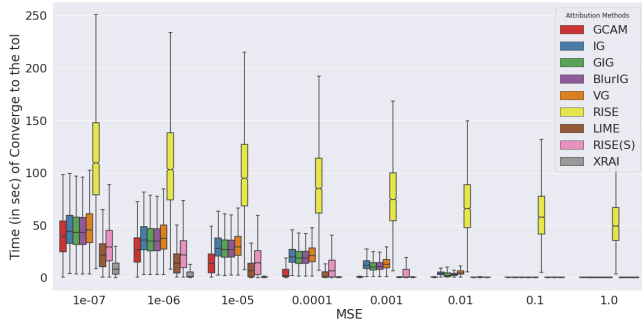}
     \vspace{-4mm}
\caption{Average value iteration converge time for all 5K test images evaluated on the \textit{InceptionV3} model.}
\label{converge_time}
\end{figure}

\subsection{Practical Analysis}
\label{practial_analysis}

\noindent \textbf{Runtime of IProp.} IProp takes 
2 minutes to construct the graph $G$ on a $299x299$ image with $K=9$. Then it takes 35 seconds to calculate the distance and apply the softmax function to generate the transition matrix $P$. The value iteration repeatedly updates the attribution map ($AM$) until convergence. Figure \ref{converge_time} shows the convergence time distribution for the value iteration on the 5K test images with the \textit{InceptionV3} model for various $tol$ and base $AM$s.

\begin{figure}[!t]
     \centering
     \begin{subfigure}[b]{\columnwidth}
         \centering
         \includegraphics[width=\textwidth]{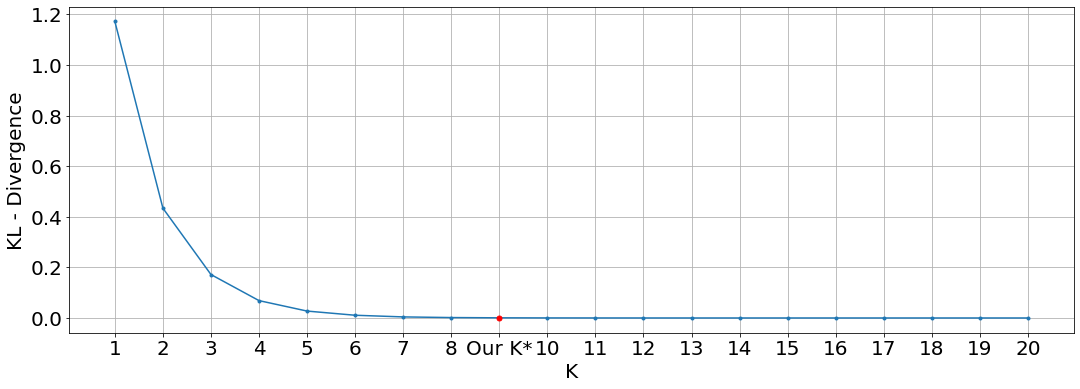}
     \end{subfigure}
\caption{KL-Divergence of two probability distribution, for $I^{th}$ pixels, generated by different $K$ when only considering the spatial distance. Each probability distribution is compared with the one generated by $K=50$.}
\label{different_k}
\end{figure}

\noindent \textbf{The Impact of Hyperparameter $\bf K$.} Intuitively, one should expect to use the $K$ that creates the connections between a pixel to all the rest of pixels, e.g., the fully connected pixel graph $G$, and let the algorithm decide the similarities between all pixel pairs. However, the fully connected graph increases the running time significantly as expected, which makes it impractical to use. On the other hand, given a pixel $I$, we observe that the similarity in the transition matrix $P$ for farther pixel $J$ is expected to have a value of zero since the geometric distance is already large enough to push the similarity to zero. We conduct an experiment where we compute the similarity vector $P^*[I,:]$ (a row in $P$) using $K=50$ for simulating dense connectivity, and use only the spatial distance as the total distance. Similarly, we compute the similarity vectors, $P^K[I,:]$, generated by different values of $K$. We hypothesize that two similarity vectors $P^*[I,:]$ and $P^K[I,:]$ will be very similar since the similarity of the pixel $I$ and further pixels is going to be zero. Furthermore, we compute the KL-divergence between $P^*[I,:]$ and $P^K[I,:]$ and present it Figure~\ref{different_k}. As the results show, and our default value of $K=9$, which is computed as image size of $299 \times 299$, and denoted as $K*$ in the figure, is a good approximation for the graph that is generated with much larger $K=50$.

\begin{figure}[!t]
     \centering
     \begin{subfigure}[b]{\columnwidth}
         \centering
         \includegraphics[width=\textwidth]{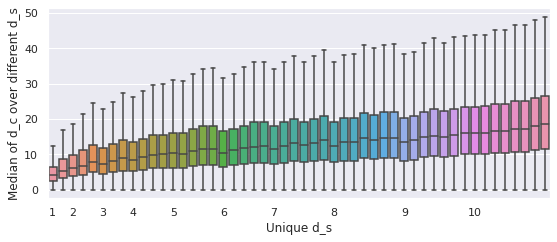}
     \end{subfigure}
\caption{Distribution of the median of $d_c$ for different unique $d_s$ over 5K test image with respect to \textit{InceptionV3}.}
\label{ds_dc_ratio}
\end{figure}

\noindent \textbf{Range for $d_s$ and $d_c$.} IProp uses both the spatial distance $d_s$ and color distance $d_c$. In this experiment, we study the range of the two distances. We calculate the spatial distance and color distance for all feasible pairs given an image. $d_c$ values are first grouped by their corresponding $d_s$. Notice that the potential unique values of $d_s$ are smaller than the number of possible neighbors $(2*K+1)^2-1$. The median value for each $d_s$ group is then recorded. Lastly, we present Fig. \ref{ds_dc_ratio}, which contains the medians for each $d_s$ over all 5K test images relative to the \textit{InceptionV3} model. As expected, the pair with larger geometric distances also has larger color distances, as distant pixel pairs are expected to be contained in distinct image objects. Note that $d_s$ and $d_c$ are within similar range, contributing equally to the overall distance.
\section{Conclusion}
We propose IProp, a novel meta-explanation method that leverages the local structural relationships of pixels and is compatible with any existing attribution map-based explanation method. IProp formulates the model explanation as an information propagation among pixels and is guaranteed to converge. Our extensive experiments show that IProp increases the explanation quality of numerous underlying explanation methods for numerous models.
In future, we plan to extend the proposed explanation approach on the graph data which have \emph{intrinsic} (causal) structure similarities~\cite{behnam2024graph}, and study the robustness of these explanation methods, as they are shown to be vulnervable in the face of adversaries~\cite{ghorbani2019interpretation,li2024graph}.  

\section* {\bf Acknowledgements} {We thank all anonymous reviewers for the constructive comments. This work of Wang was supported by Wang's startup funding, the Cisco Research Award, and the National Science
Foundation under grant No.~2216926,~2241713,~2331302, and~2339686.}
\newpage
\bibliographystyle{ACM-Reference-Format}
\bibliography{reference}

\appendix

\end{document}